\documentclass[letterpaper, 10 pt, conference]{ieeeconf}  

\IEEEoverridecommandlockouts                      

\usepackage{epsfig} 
\usepackage{amsmath} 
\usepackage{amssymb}  
\usepackage{color}
\usepackage[linesnumbered,ruled]{algorithm2e}
\usepackage[normalem]{ulem} 
\makeatletter
\let\NAT@parse\undefined
\makeatother
\usepackage{bbding}
\usepackage{hyperref}
\hypersetup{pdfstartview=FitH,
            colorlinks=true,
            linkcolor=red,
            anchorcolor=blue,
            citecolor=green
            }
\usepackage{cite}
\usepackage{booktabs}
\usepackage{multirow}
\usepackage[caption=false,font=footnotesize]{subfig}
\usepackage[dvipsnames]{xcolor}
\usepackage{subfig}
\usepackage{soul}
\usepackage{balance}
\usepackage{stfloats}
\newcounter{RNum}
\renewcommand{\theRNum}{\arabic{RNum}}
\newcommand{\Remark}{\noindent\textit{\textbf{Remark}~\refstepcounter{RNum}\textbf{\theRNum}: }}
\newcommand{\NoOne}[1]{\textcolor{red}{#1}}
\newcommand{\NoTwo}[1]{\textcolor{green}{#1}}
\newcommand{\NoThree}[1]{\textcolor{blue}{#1}}
\soulregister\NoOne7
\soulregister\NoTwo7
\soulregister\NoThree7
\soulregister\Remark7
\definecolor{table_c}{RGB}{245,245,245}
\title{\LARGE \bf
EdgeSpotter: Multi-Scale Dense Text Spotting for Industrial Panel Monitoring
}
\author{Changhong Fu$^{1, 2, *}$, Hua Lin$^{1}$, Haobo Zuo$^{3}$, Liangliang Yao$^{1}$, Liguo Zhang$^{1, 4}$ %
\thanks{$^{*}$Corresponding author}
\thanks{
$^{1}$C. Fu, H. Lin, L. Yao, and L. Zhang are with the School of Mechanical Engineering, Tongji University, Shanghai 201804, China.
\itshape{Email: changhongfu@tongji.edu.cn}
}
\thanks{$^{2}$C. Fu is with the Shanghai Key Laboratory of Wearable Robotics and Human-Machine Interaction, Tongji University, Shanghai 201804, China.} 
\thanks{
$^{3}$H. Zuo is with the Department of Computer Science, University of Hong Kong, Hong Kong 999077, China.
}
\thanks{
$^{4}$L. Zhang is with the Department of NANO Fabrication, Institute of Nano-Tech and Nano-Bionics (SINANO), Chinese Academy of Sciences, Suzhou 215123, China.
}
}
\begin{document}
\maketitle
\thispagestyle{empty}
\pagestyle{empty}
\begin{abstract} Text spotting for industrial panels is a key task for intelligent monitoring. However, achieving efficient and accurate text spotting for complex industrial panels remains challenging due to issues such as cross-scale localization and ambiguous boundaries in dense text regions. Moreover, most existing methods primarily focus on representing a single text shape, neglecting a comprehensive exploration of multi-scale feature information across different texts. To address these issues, this work proposes a novel multi-scale dense text spotter for edge AI-based vision system (EdgeSpotter) to achieve accurate and robust industrial panel monitoring. Specifically, a novel Transformer with efficient mixer is developed to learn the interdependencies among multi-level features, integrating multi-layer spatial and semantic cues. In addition, a new feature sampling with catmull-rom splines is designed, which explicitly encodes the shape, position, and semantic information of text, thereby alleviating missed detections and reducing recognition errors caused by multi-scale or dense text regions. Furthermore, a new benchmark dataset for industrial panel monitoring (IPM) is constructed. Extensive qualitative and quantitative evaluations on this challenging benchmark dataset validate the superior performance of the proposed method in different challenging panel monitoring tasks. Finally, practical tests based on the self-designed edge AI-based vision system demonstrate the practicality of the method. The code and demo will be available at \url{https://github.com/vision4robotics/EdgeSpotter}. 

\end{abstract}
\section{Introduction} \label{sec:intro}
Intelligent vision systems play a vital role in industrial automation \cite{gao2022review}, \cite{shang2024methods} especially for the field of industrial panel monitoring. Industrial panels, including oscilloscopes, temperature controllers, aging test platforms, etc., can provide real-time data. However, limited access to this data hampers effective management and complicates subsequent verification. Currently, industrial panel data monitoring largely relies on manual regular observation and record-keeping. This approach not only fails to capture the full lifecycle data of product manufacturing and testing but also risks the loss of critical data and inaccurate performance evaluations due to the high-intensity manual labor required. Furthermore, with the rapid evolution of equipment and the continuous growth of the electronics industry \cite{jiang2024comprehensive}, traditional manual monitoring can no longer meet the increasing demands for efficiency, quality, and reliability in modern industrial production. The need for continuous (24/7) monitoring has increased, highlighting the demand for intelligent monitoring solutions for panel equipment. \textit{\textbf{Therefore, the design of an intelligent vision system with high accuracy, efficiency, and robustness for industrial panel monitoring tasks has become an urgent priority.}}\par
\begin{figure}[!t]
	\centering
	\includegraphics[width=1.0\linewidth]{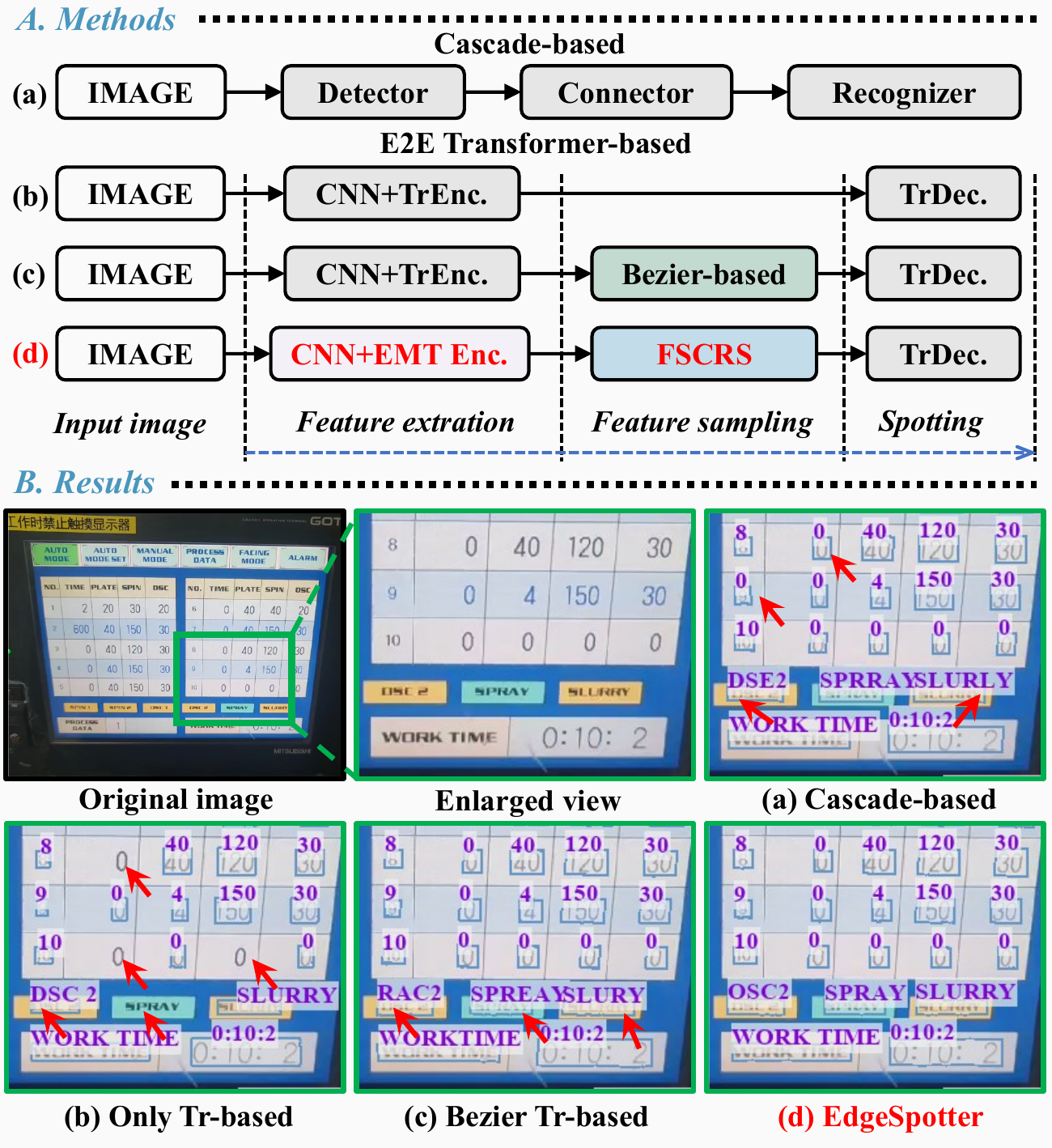} 
	\setlength{\abovecaptionskip}{-5pt} 
	\caption
    {An overview of state-of-the-art (SOTA) text spotting methods and the proposed EdgeSpotter. \textbf{(a)} A cascade approach with separate text detection and recognition stages. \textbf{(b)} A Transformer-based framework for concurrent optimization of detection and recognition. \textbf{(c)} Bezier-based feature sampling integrated into the Transformer framework. \textbf{(d)} Efficient multi-scale feature extraction and a novel feature sampling for performance enhancement. TrEnc.: Transformer encoder. TrDec.: Transformer decoder. Tr-Based: Transformer-based. EMT Enc.: Encoder with efficient mixer transformer. E2E: end-to-end. Red arrows indicate incorrect spotting results.
	}
	\label{fig:fig1}
        \vspace{-10pt}
\end{figure}
Deep learning-based text detection and recognition technologies have emerged as innovative solutions for industrial automation monitoring \cite{deng2023end}, \cite{park2024ocr}. Due to their successful applications in various domains, cascade-based text spotting methods \cite{wang2021pan++}, \cite{liu2020abcnet}, \cite{liu2021abcnet} have been extensively validated. They follow a detect-then-recognize strategy, as shown in Fig. \ref{fig:fig1}. However, the cascade structure inherently leads to performance degradation from the detection network to the recognition network. Additionally, under this paradigm, there is a lack of synergy between text detection and recognition. These limitations hinder the wider adoption of text spotting technologies in complex industrial monitoring scenarios, such as those involving dense text, multi-scale variations, and strong lighting conditions.\par
In recent years, with the widespread application of Transformers in computer vision \cite{vaswani2017attention}, \cite{han2022survey}, researchers have increasingly integrated detection and recognition networks into Transformer frameworks to enhance feature interaction. These end-to-end (E2E) Transformer-based text spotting methods \cite{huang2022swintextspotter}, \cite{ye2023deepsolo}, \cite{qiao2024dntextspotter} have garnered significant attention for their ability to leverage the synergy between text detection and recognition. However, despite their satisfactory performance in general scenarios, these methods often overlook unique challenges posed by tasks like text scale variation, dense regions, and complex conditions in industrial panel monitoring. Moreover, research specifically targeting industrial applications is still limited compared to cascade-based methods \cite{wang2021pan++}, \cite{liu2020abcnet}, \cite{liu2021abcnet}. Furthermore, the practical effectiveness of these methods has not been thoroughly validated in real-world industrial settings, and their high computational demands present challenges for edge deployment. \textit{\textbf{Therefore, it is crucial to develop a lightweight Transformer-based text spotter that is both efficient and robust, tailored to address the specific challenges of industrial panel text spotting.}}\par
Due to the significant scale variations among text objects in panels, single-layer feature-based Transformers struggle to effectively represent multi-scale information. To address this, this work designs an efficient multi-level feature mixer to learn the interdependencies among features at different levels, thereby adapting to text of varying scales. Additionally, to tackle the issue of inaccurate spotting in dense text regions, this work introduces feature sampling based on catmull-rom splines, which explicitly encodes the shape, position, and semantic information of text. Overall, a novel multi-scale dense text spotter for edge AI-based vision systems (EdgeSpotter) is proposed to achieve accurate and robust industrial panel monitoring. The main contributions of this work are as follows:
\begin{itemize}
\item An edge intelligence-driven spotting framework for industrial panel monitoring is designed and implemented in the self-built edge AI system, enabling continuous (24/7) monitoring of panel information compared to the burdensome, labor-intensive and short-term nature of manual monitoring.
\item An efficient multi-level feature mixer is proposed to learn the interdependencies and spatial information among different features, thereby enhancing text spotting performance at arbitrary scales.
\item Feature sampling based on catmull-rom splines is introduced to mitigate the boundary ambiguities and recognition errors caused by dense text regions. 
\item A text dataset tailored for industrial panel monitoring scenarios (\textbf{IPM}) is meticulously constructed. Extensive experiments on this dataset demonstrate the superior performance and robustness of EdgeSpotter. Furthermore, real-world deployment applications further validate its effectiveness in practical scenarios. \end{itemize}

\section{Related Work}
\subsection{Text Spotting}
In the past, text spotting was typically divided into two distinct subtasks: detection and recognition \cite{ren2016faster},\cite{zhou2017east}, \cite{liao2020real}, \cite{cheng2018aon} with each task being studied independently. However, these methods exhibit several limitations, such as inefficient inference of dense characters, error accumulation, and suboptimal performance. To alleviate these issues, text spotting methods have gradually evolved from a two-stage shallow learning framework to an E2E deep learning architecture. Initially, H. Li \textit{et al.} \cite{li2017towards} integrated detection and recognition into a unified network, enabling joint optimization of both tasks. Nevertheless, this method is limited to handling horizontal text. To accommodate oriented text, researchers introduce specialized region-of-interest (RoI) operations, such as Text-Align and RoI-Rotate \cite{he2018end}, which transform oriented text features into regular features for recognition. Mask TextSpotter \cite{lyu2018mask} tackles arbitrary-shaped text recognition by incorporating a character segmentation module that leverages character-level annotations. Furthermore, TextDragon \cite{feng2019textdragon} introduces RoISlide, which combines features from predicted segments, thereby enhancing text recognition. As research has advanced, methods such as Mask TextSpotter V3 \cite{liao2020mask}, MANGO \cite{qiao2021mango} and ABCNet \cite{liu2020abcnet} have contributed to the shift toward E2E deep learning frameworks. These methods emphasize the crucial synergy between text detection and recognition. However, most of them achieve task collaboration through a shared backbone network. In essence, they still follow the detection-then-recognition paradigm, and the underlying challenges remain, particularly in complex tasks such as panel text spotting.\par
\begin{figure*}[!t]	
	\centering
	\includegraphics[width=1\linewidth]{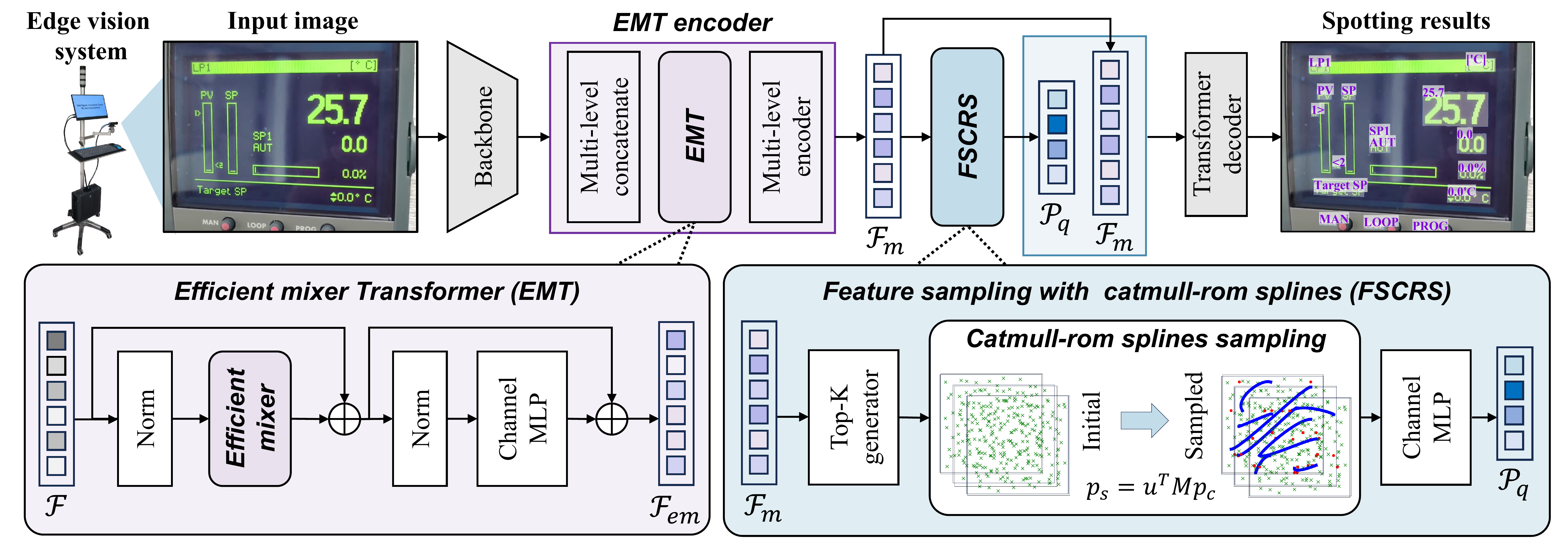}
        \setlength{\abovecaptionskip}{-5pt} 
        \vspace{-0.3cm}
	\caption
        {An overview of the proposed EdgeSpotter. The modules from the left to right are \emph{backbone}, \emph{encoder with efficient mixer transformer (EMT encoder)}, \emph{feature sampling with catmull-rom splines (FSCRS)} and \emph{transformer decoder}. The last three feature maps extracted by backbone are fed to \emph{EMT encoder}.}
	\label{fig:fig2}
        \vspace{-10pt}
\end{figure*}
With the proven effectiveness of Transformers in computer vision tasks \cite{han2022survey}, researchers have successfully integrated detection and recognition networks into Transformer frameworks to enhance feature interaction between text detection and recognition tasks. TETSR \cite{zhang2022text} adopts a dual-decoder framework with a shared backbone and encoder, while providing separate heads for detection and recognition tasks. SwinTextSpotter \cite{huang2022swintextspotter} introduces a recognition transformation mechanism that implicitly guides the recognition head by merging detection information and recognition feedback into the detector. While these methods eliminate the need for heuristic post-processing, they still face challenges related to task collaboration and training efficiency. To address these issues, DeepSolo \cite{ye2023deepsolo} and DNtextSpotter \cite{qiao2024dntextspotter} explicitly model a set of learnable point sequences, deriving text centerlines, boundaries, instances, and confidence from a single decoder. Moreover, most of the existing work is primarily focused on text spotting in natural scenes and lacks robustness when dealing with complex text information, especially with significant scale variations in industrial panels. Additionally, the large computational demands of these methods lead to real-time performance issues, which are critical for industrial applications. To address these challenges, this work designs a novel efficient mixer transformer and implements multi-level feature mixing through a scale-adaptive vector, effectively aggregating multi-scale text information for precise and robust panel text spotting.\par
\subsection{Feature Sampling}
Initially, in cascade-based methods, feature sampling served as an important mechanism for connecting detector and recognizer, providing the raw input for recognition operations from potential text areas. RoI Pooling \cite{girshick2015fast} was first introduced for feature sampling and has since been widely adopted \cite{li2017towards}, \cite{litman2020scatter}. RoIAlign \cite{he2017mask} employs bilinear interpolation for weighted feature sampling and was the first to be extended to sample non-axis-aligned (\textit{i.e.}, rotated) RoIs. Notably, ABCNet \cite{liu2020abcnet} introduces Bezier parameterization to localize curved text, performing feature sampling through a Bezier alignment operator, which effectively addresses the issue of detecting arbitrarily-shaped text. More complexly, GLASS \cite{ronen2022glass} integrates additional information calculated directly from normalized word crops, enabling feature sampling from global to local scales. With the advent of subsequent E2E Transformer methods, feature sampling typically appears between the encoder and decoder to extract text features. For instance, TESTR \cite{zhang2022text} acquires the raw control point sequence via the Guidance Generator and then uses the Box-to-Polygon strategy to obtain reference features for the subsequent decoder. Although it uses point queries with bounding box positions, the queries for detection and recognition are distinct. Similarly, inspired by ABCNet \cite{liu2020abcnet}, DeepSolo \cite{ye2023deepsolo} utilizes center Bezier curve sampling to encode text features explicitly. However, the control points in the aforementioned methods are highly indirect with respect to the sampled curves, making them less suitable for text with relatively regular shapes, especially in the case of panel text.

\section{Proposed Method}
The workflow of EdgeSpotter is presented in Fig. \ref{fig:fig2}. It can be divided into four modules: \textit{backbone}, \textit{encoder with efficient mixer Transformer}, \textit{feature sampling with catmull-rom splines}, \textit{Transformer decoder}. To ensure consistency with other SOTA methods, this work adopts ResNet50 as the backbone for feature extraction.

\subsection{Encoder with Efficient Mixer Transformer (EMT Encoder)}
To fully explore the interdependencies among multi-level features, EdgeSpotter utilizes the output features from the last tree stages of the backbone. Specifically, the last three feature maps are fed into efficient mixer Transformer (EMT). Through efficient mixer (EM), intra-scale interaction and cross-scale fusion are realized to obtain multi-scale features with rich information. The simple network structure ensures the operation efficiency.\par 
For clarity, the last three output feature maps are uniformly represented by $\mathcal{F}_l\in{\mathbb{R}}^{W \times H \times C}$ in the following introduction ($C$, $W$ and $H$ represent the channel, width, and height of the feature maps respectively, and $l \in \{3, 4, 5 \}$). Then,  $\mathcal{F}_5$ undergoes $3 \times3$ Conv with an additional step size of 2 to obtain $\mathcal{F}_6$. Finally, $\mathcal{F}_3$, $\mathcal{F}_4$, $\mathcal{F}_5$ and $\mathcal{F}_6$ are reshaped and concatenated to obtain the final feature $\mathcal{F} \in {\mathbb{R}}^{N \times C}$, where $N$ is the token length. Then, EMT can be expressed as follows:
\begin{align}
\mathcal{{X}} &= \text{EM}(\text{LN}(\mathcal{F}))+\mathcal{F}) \ \ \text{,}\\
\text{EMT}(\mathcal{F}) &= \text{MLP}(\text{LN}(\mathcal{X}))+\mathcal{X} \ \ \text{,}
\end{align}
where $\text{LN}(\cdot)$ is layer normalization, MLP ($\cdot$) is multi-layer perception and $\text{EM}(\cdot)$ represents the efficient mixer.\par

\noindent \textbf{\textit{Remark 1:}} EMT emphasizes the interaction among multi-level features, facilitating the construction of inter-layer dependencies and enabling multi-level feature complementarity. Subsequently, a multi-level encoder is applied using deformable Transformer.\par

\begin{figure}[!t]	
    \vspace{5pt}
    \centering
    \includegraphics[width=1\linewidth]{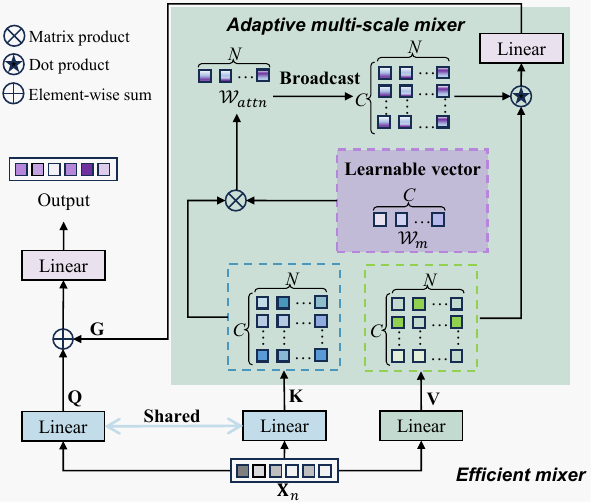}
    \setlength{\abovecaptionskip}{-5pt} 
    \caption
    {
   Detailed workflow of efficient mixer. The input features are processed through adaptive multi-scale mixer after linear mapping. Here, the learnable vector $\mathcal{W}_{m}$ is used to learn the attention weights for the query, and the generated attention map $\mathcal{W}_{attn}$ is multiplied with the principal elements of matrix V via a broadcast mechanism.
    }
    \label{fig:fig3}
    \vspace{-10pt}
\end{figure}

Inspired by \cite{shaker2023swiftformer}, EM avoids expensive matrix multiplication operations and reduces the computational complexity from quadratic to linear. At the same time, a learnable parameter vector $\mathcal{W}_m\in \mathbb{R}^{C}$ is introduced to represent the multi-level attention weights. Since the obtained feature  $\mathcal{F}$ contains rich multi-level features, $\mathcal{W}_m$ is used to reshape the specific type of text information (such as different scales) in each layer. This not only avoids the information confusion caused by the interaction among different levels, but also facilitates the dynamic exploration of multi-level feature information. Specifically, the input embedding matrix $\textbf{X}_n$ is transformed into $\textbf{Q}$, $\textbf{K}$ and $\textbf{V} \in \mathbb{R}^{N \times C}$ using two matrices $\textbf{W}_k$, $\textbf{W}_v$ ($\textbf{K}$ is equal to $\textbf{Q}$). Next, the matrix $\textbf{K}$ is multiplied by $\mathcal{W}_{m}$ to learn the attention weights of the query, producing global attention query vector $\mathcal{W}_{attn} \in \mathbb{R}^{N}$. Subsequently, the matrix $\textbf{V}$ is element-wise multiplied with the broadcasted  $\mathcal{W}_{attn}$ to yield the global context representation. This operation integrates global information into each element of the matrix, thereby enhancing the model's sensitivity to multi-scale features. After one more linear layer, the resulting multi-scale attention matrix $\textbf{G} \in \mathbb{R}^{N \times C}$  is added element-wise with \textbf{Q}. As shown in Fig. \ref{fig:fig3}, EM can be summarized as:
\begin{align}
    \mathcal{W}_{attn} &= \textbf{K}  \mathcal{W}_{m} \ \ \text{,} \\
    \text{EM}(\textbf{X}_n)&=\varphi(\varphi( \textbf{V} \star \mathcal{W}_{attn} / \sqrt{\text{D}})\oplus \textbf{Q}) \ \ \text{,}
\end{align}
where $\textbf{X}_n$ is the input feature, $\star$ represents dot product, and $\varphi(\cdot)$ indicates the linear layer. The proposed matrix operations capture information from each token and learn correlations within the input sequence.\par

\noindent \textbf{\textit{Remark 2:}} EM employs element-wise multiplication to achieve linear complexity, which greatly reduces the computational cost. In addition, $ \mathcal{W}_{m}$ adaptively learns the attention weights of multi-level features, so that the model has strong robustness when dealing with arbitrary scale text.\par

\subsection{Feature Sampling with  Catmull-Rom Splines (FSCRS)}
Common feature sampling methods, such as box selection proposals \cite{zhang2022text} and Bezier curve proposals \cite{ye2023deepsolo}, either exhibit limitations in information representation or suffer from poor smoothness due to the indirect relationship between the shape of a curve and its control points. In contrast, this work proposes a simple scheme based on catmull-rom splines, named feature sampling with catmull-rom splines (FSCRS), which is designed from the text center-line perspective. FSCRS efficiently fits panel text and distinguishes between different text instances. This feature sampling method is then used to guide subsequent feature decoding, effectively alleviating boundary ambiguities caused by dense panel text. \par
\begin{figure}[!t]	
    \vspace{3pt}
    \centering
    \includegraphics[width=1\linewidth]{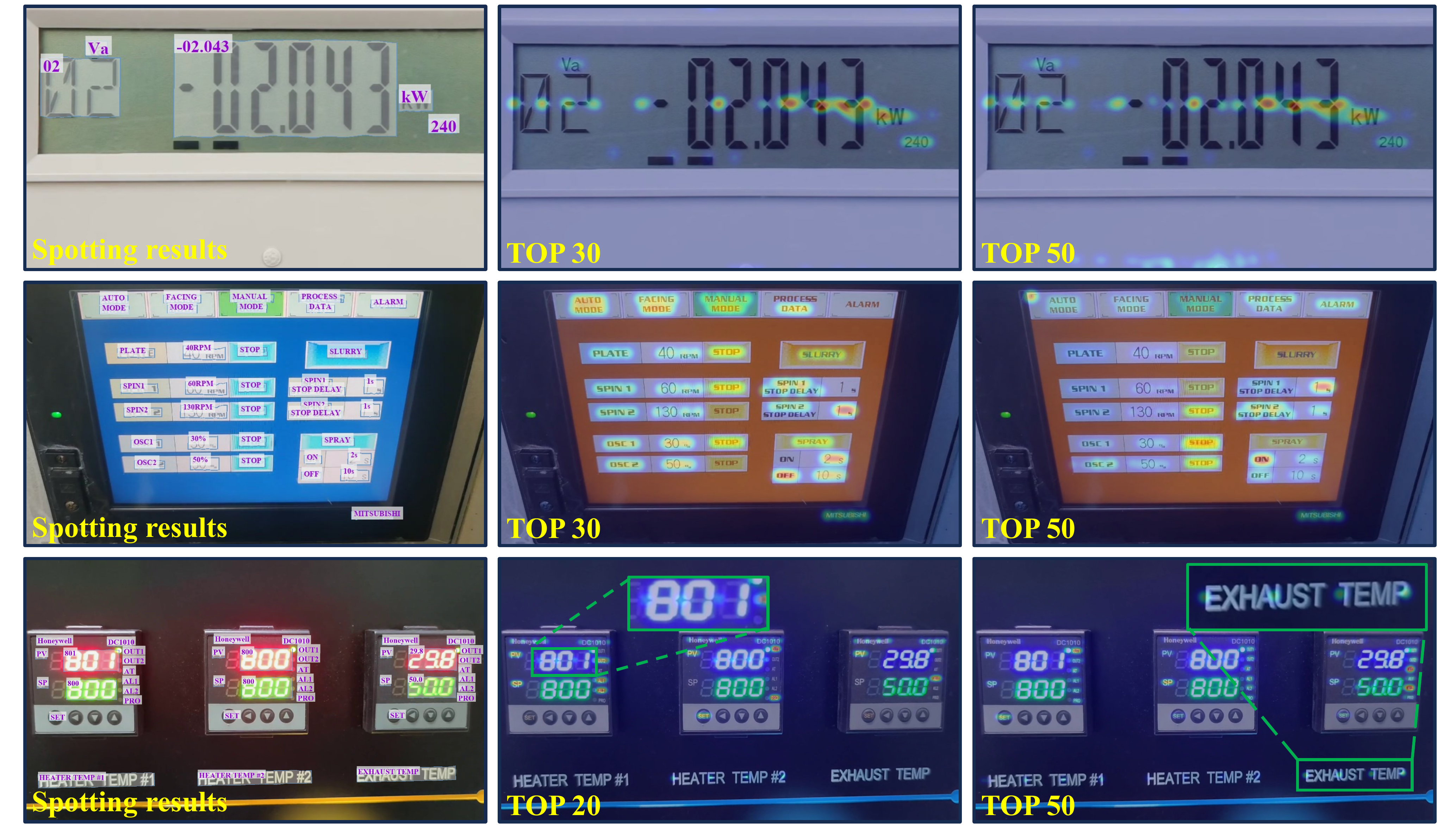}
    \setlength{\abovecaptionskip}{-5pt} 
    \caption
    {
   Density maps of control points for Top-K scores. Text instances with lower scores tend to have a greater number of control points. Zoom in for better visualization.
    }
    \label{fig:fig4}
    \vspace{-10pt}
\end{figure}

Given the image features from the feature extraction network, on each pixel of the features, a MLP is used to predict offsets to four catmull-rom control points, determining a curve that represents one text instance. Let \( i \) index a pixel from the features, with 2D normalized coordinates \( \hat{p}_i = (\hat{p}_{ix}, \hat{p}_{iy}) \in [0,1]^2 \). For each pixel \( i \), the corresponding catmull-rom control points \( \textbf{CRP}_i = \{crp_{i_0}, crp_{i_1}, crp_{i_2}, crp_{i_3}\} \) are predicted. The coordinates of these control points are computed using the sigmoid function \( \sigma \) as follows: 
\begin{equation}\label{eqn-3}
    \begin{aligned}
crp_{i_j} &= \sigma(\Delta p_{i_j} + \sigma^{-1} (\hat{p}_{i}))\ \ \text{,}\\
    \end{aligned}
\end{equation}
where $j \in \{0, 1, 2, 3\}$. Control points with top-$K$ are selected as the proposals based on the scoring results of a linear layer. The initial control point queries described are denoted as $\{\textbf{P}_{c}^{(k)}\}^{K}_{k=1}$. Then, $n$ points on each curve are uniformly sampled \cite{twigg2003catmull} to finally obtain the point coordinates as $\textbf{P}_s \in \mathbb{R}^{K \times n \times 2}$. Based on $\textbf{P}_s$, we use a simple MLP for further projection to get the positional queries $\textbf{P}_q \in \mathbb{R}^{K \times n \times C}$. The computation of $\textbf{P}_q$ can be formulated as follows:
\begin{align}
    \textbf{P}_s^{(k)}&=\text{CatRom}(\textbf{P}_{c}^{(k)})\quad, \\
    \textbf{P}_q&=\text{MLP}(\text{PE}(\textbf{P}_s))\quad ,
\end{align}
where CatRom$(\cdot)$ denotes catmull-rom splines sampling and $\text{PE}(\cdot)$ represents the sinusoidal positional encoding function. CatRom$(\cdot)$ can be formulated as follows:
\begin{equation}\label{eqn-5}
    \begin{aligned}
    \text{CatRom}(\textbf{P}_{c}^{(k)})=\textbf{U}(u)\textbf{M}(\tau)\textbf{P}_{c}^{(k)} \ \ \text{,}
    \end{aligned}
\end{equation}
where $u\in[0, 1]$ represents the curve parameters and $\tau$ is the tension parameter used to control the smoothness of the curve ($\tau=0.5$ is set empirically). The coefficients of the matrix $\textbf{M}$ determine the shape of the catmull-rom splines. Even in challenging panel text monitoring scenarios, such as dense and multi-scale text, FSCRS can still accurately locate all text instances and extract key features, as demonstrated in Fig. \ref{fig:fig4}.\par
\noindent \textbf{\textit{Remark 3:}} By analyzing the density distribution of the Top-K control points, it is observed that text instances with lower scores tend to have a greater number of control points, as shown in Fig. \ref{fig:fig4}. This indicates that FSCRS can dynamically adjust the number of control points for different text instances based on their scores, leading to an overall improvement in performance.\par
\subsection{Transformer Decoder}
As shown in Fig. \ref{fig:fig2}, the feature queries obtained by FSCRS is fed into the subsequent decoder as reference points, while the multi-scale features extracted by the EMT encoder serve as the input to the decoder, forming the base sequence. Then, following \cite{ye2023deepsolo}, this work employs four simple prediction heads to predict the instance class, character class, center curve points, and bounding box, respectively.\par
\begin{table*}[bp]
    \centering
    \caption{Quantitative spotting results on the benchmark. All evaluations were performed on a single NVIDIA TITAN RTX. Red represents the best result.}
    \setlength{\tabcolsep}{4.8mm}
    \renewcommand{\arraystretch}{1.1}
    \begin{tabular}{c|c|c|c|c|c|c|c|c}
        \toprule[0.5mm]
        \multirow{2}*{\textbf{Method}} & \multirow{2}*{\textbf{Years}} & \multirow{2}*{\textbf{Backbone}} & \multicolumn{3}{c|}{\textbf{Detection}} & \multicolumn{3}{c}{\textbf{Recognition}} \\
        \cline{4-9}
        & & & \textbf{P($\%$)}& \textbf{R($\%$)} &  \textbf{F1($\%$)} & \textbf{P($\%$)} & \textbf{R($\%$)} & \textbf{H($\%$)}\\
        \midrule[0.25mm]
        ABCNet \cite{liu2020abcnet} &2020 & ResNet50 & 97.29 & 90.39 & 93.71 & 74.43 & 69.16 & 71.70\\
        Mask TexSpotter v3 \cite{liao2020mask} &2020 & ResNet50 & 92.54 & 89.67 & 91.08 & 73.26 & 69.34 & 71.25 \\
        PGNet \cite{wang2021pgnet} &2021 & ResNet50  & 68.00 & 63.38 & 65.61 & 43.60 & 35.74 & 39.25 \\
        ABCNetv2 \cite{liu2021abcnet} &2021 & ResNet50  & 95.02 & 91.90 & 93.43 & 73.09 & 71.22 & 72.14 \\
        PAN++ \cite{wang2021pan++} &2021 & ResNet18 & 93.26 & 88.15 & 90.63 & 73.15 & 70.69 & 71.89  \\
        SwinTextSpotter \cite{huang2022swintextspotter} &2022 & Swin-T  & 96.11 & 90.57 & 93.26 & 80.31 & 75.69 & 77.93 \\
        DeepSolo \cite{ye2023deepsolo} &2023 & ResNet50  & 97.67 & 93.78 & 95.68 & 78.60 & 75.47 & 77.00\\
        DNtextSpotter \cite{qiao2024dntextspotter} &2024 & ResNet50  & 93.21 & 94.23 & 93.72 & 78.12 & 76.35 &77.22  \\
        \midrule
        \textbf{EdgeSpotter (Ours)} &- & ResNet50  & \textbf{\textcolor{red}{98.16}} & \textbf{\textcolor{red}{94.52}} & \textbf{\textcolor{red}{96.31}} & \textbf{\textcolor{red}{82.53}} & \textbf{\textcolor{red}{80.02}} & \textbf{\textcolor{red}{81.25}} \\
        \bottomrule[0.5mm]
    \end{tabular}
    \label{tab:tab1}
\end{table*}
\noindent \textbf{\textit{Remark 4:}} Through EMT, the spatial and semantic information across multi-level features is fully leveraged, enhancing robustness when handling large-scale variations. Simultaneously, FSCRS is employed to extract critical information from these multi-level features, which significantly benefits the final feature decoding. With just four parallel, simple prediction heads, EdgeSpotter achieves satisfactory results, as shown in Fig. \ref{fig:fig5}.\par
\section{Experiments}
\subsection{Datasets}
This work evaluates the performance of our method on a new benchmark dataset, \textbf{IPM}, which is specifically designed for industrial panel monitoring and contains 2,005 images. Of these, 1,200 images are used for training, 300 images for validation, and the remaining images are classified according to various challenge attributes. These attributes include 'side view', 'strong light', 'reflection', 'shadow', 'special character', 'multi-scale', and 'dense character'. Additionally, this work uses the following datasets for pre-training: 1) \textbf{Synth150K} \cite{liu2020abcnet}, a synthetic dataset containing 94,723 images with multi-oriented text and 54,327 images with curved text; 2) \textbf{IC15} \cite{karatzas2015icdar}, which includes 1,000 training images and 500 test images for quadrilateral scene text; and 3) \textbf{CTW1500} \cite{liu2019curved}, a text-line level benchmark for scene text with arbitrary shapes, consisting of 1,000 training images and 500 test images.\par
\noindent \textbf{\textit{Remark 5:}} To ensure better generalization and efficient convergence, pre-training on diverse datasets is essential. \textbf{IPM} is then used for fine-tuning, encompassing a variety of industrial panel types and specifically designed for industrial panel monitoring. The dataset includes 31,595 text targets, with an average of 16 targets per image.
\begin{figure*}[!t]	
	\centering
	\includegraphics[width=1\linewidth]{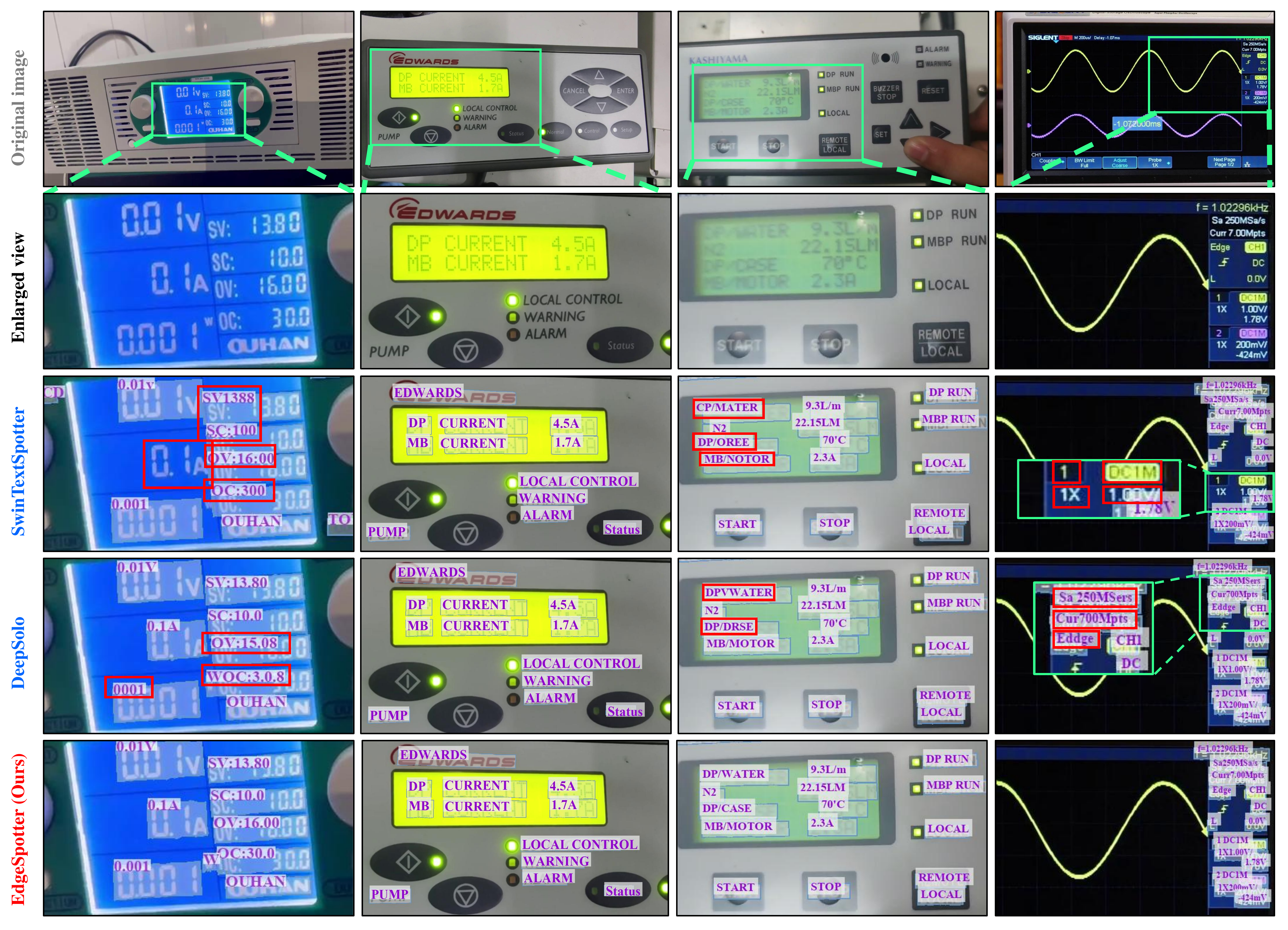}
        \setlength{\abovecaptionskip}{-5pt} 
	\caption
        {Comparison of EdgeSpotter spotting results with other SOTA spotting results. Areas with incorrect identification (\textit{e.g., incorrect identification, missed detection)} are marked with red boxes. The original image and the enlarged view can be found in the second and first rows, respectively. }
	\label{fig:fig5}
        \vspace{-10pt}
\end{figure*}
\subsection{Implementation Details and Evaluation Metric}

\subsubsection{Implementation Details}
In our model, the number of proposals \(K\) is set to 100, while the number of sampled points \(n\) is set to 25. EdgeSpotter predict 96 classes on \textbf{IPM}. The AdamW optimizer is utilized for training. Model training and evaluation were performed on the NVIDIA TITAN RTX GPUs. To be fair, all models were trained for a uniform 12k iterations (cascade-based methods may be trained for more iterations to ensure results).
\subsubsection{ Evaluation Metric}The evaluation protocols used in the ICDAR Robust Reading Competition are applied to assess detection performance \cite{karatzas2015icdar}. Specifically, a bounding box is considered correct if its IoU with any ground truth exceeds 0.5, and the recognized word also matches. For recognition, this work follows the "End-to-End" evaluation protocol \cite{li2017towards}, which requires all words in the image to be recognized, regardless of whether the string exists in the provided contextual lexicon.

\subsection{Comparison with State-of-the-art}
\subsubsection{Overall Performance}To comprehensively evaluate the effectiveness of the proposed method, this work compares our model with eight SOTA text spotting methods on the IPM validation set, including ABCNet series \cite{liu2020abcnet}, \cite{liu2021abcnet}, Mask TextSpotter v3 \cite{liao2020mask}, PGNet \cite{wang2021pgnet}, PAN++ \cite{wang2021pan++}, SwinTextSpotter \cite{huang2022swintextspotter}, DeepSolo \cite{ye2023deepsolo}, and DNtextSpotter \cite{qiao2024dntextspotter}. For fairness, all methods use ResNet50 as the backbone network wherever possible. As shown in TABLE \ref{tab:tab1}, EdgeSpotter outperforms other SOTA text spotting methods in terms of overall performance. Specifically, EdgeSpotter achieves the highest F1 and H, with a \textbf{0.63\%} and \textbf{3.32\%} improvement over the second-best method on these two metrics, respectively. Fig. \ref{fig:fig5} shows some visualization examples. The superior performance in the comprehensive evaluation demonstrates that EdgeSpotter is the optimal choice for industrial panel monitoring.
\subsubsection{Attribute-based Comparison}To thoroughly evaluate EdgeSpotter under various challenges, an attribute-based comparison is conducted, as shown in Fig. \ref{fig:fig6}. Based on the differences in methodology and the comprehensive performance data presented in the TABLE~\ref{tab:tab1}, five methods are compared in this section. Overall, EdgeSpotter ranks first in both detection and recognition performance. Specifically, it significantly outperforms the second-best methods in attributes such as multi-scale, dense text, strong light, and side view. Among these, the most notable improvement is seen in the detection of dense text, while the most substantial enhancement in recognition is observed for multi-scale text. The satisfactory results demonstrate that the efficient multi-level feature mixing strategy and the feature sampling method based on Catmull-Rom splines, can effectively improve the performance of industrial panel text spotting, particularly in complex scenes, including multi-scale and dense scenarios.
\setcounter{figure}{6}
\begin{figure*}[!t]	
	\centering
	\includegraphics[width=1\linewidth]{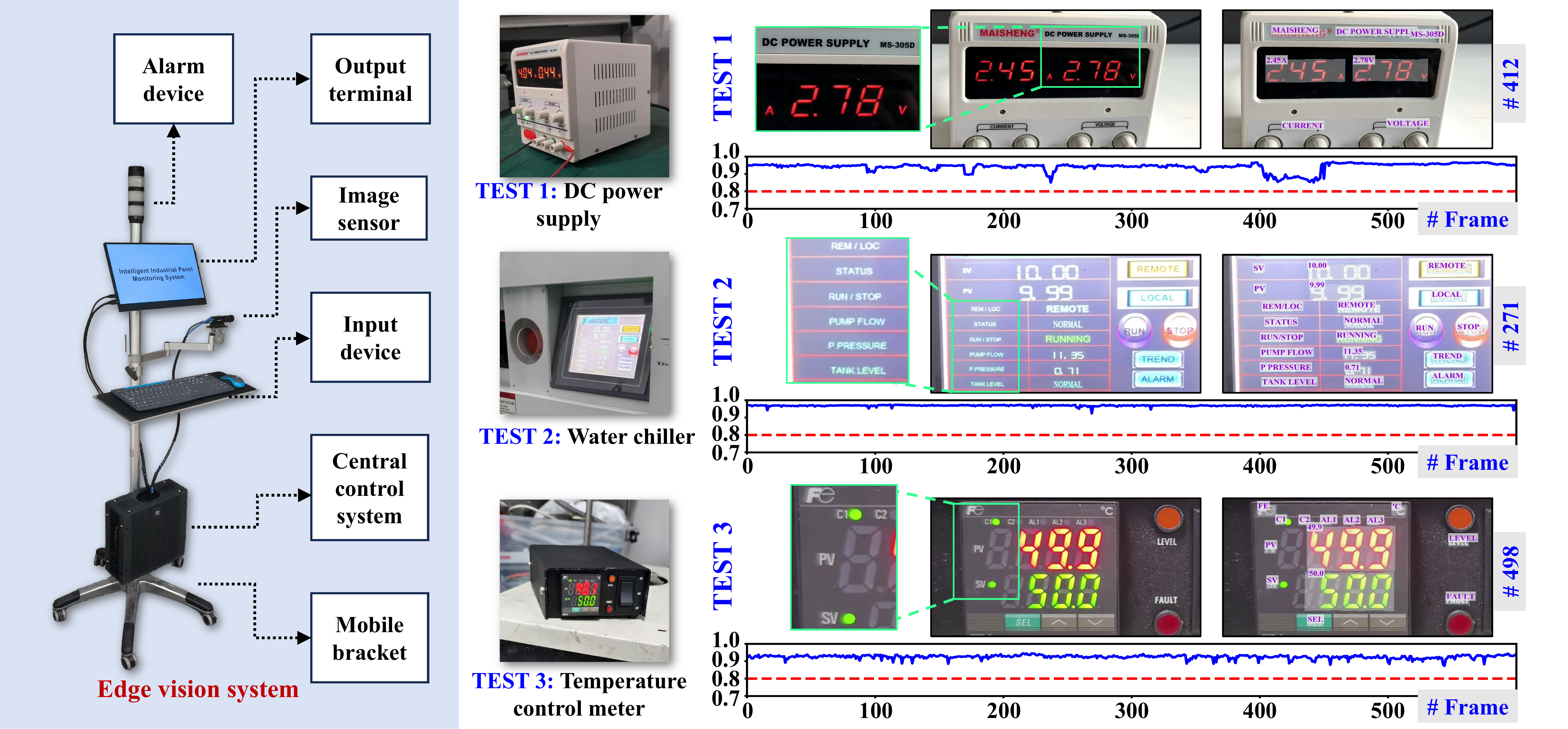}
        \setlength{\abovecaptionskip}{-5pt} 
	\caption
        {Visualization of real-world tests. In real-world applications, scores above the \textcolor{red}{red} dashed line are considered accurate and reliable spotting results. The \textcolor{blue}{blue} solid line represents the actual test scores. Zoom in for better visualization.}
	\label{fig:fig7}
        \vspace{-10pt}
\end{figure*}
\setcounter{figure}{5}
\begin{figure}[t]	
    \vspace{5pt}
    \centering
    \includegraphics[width=1\linewidth]{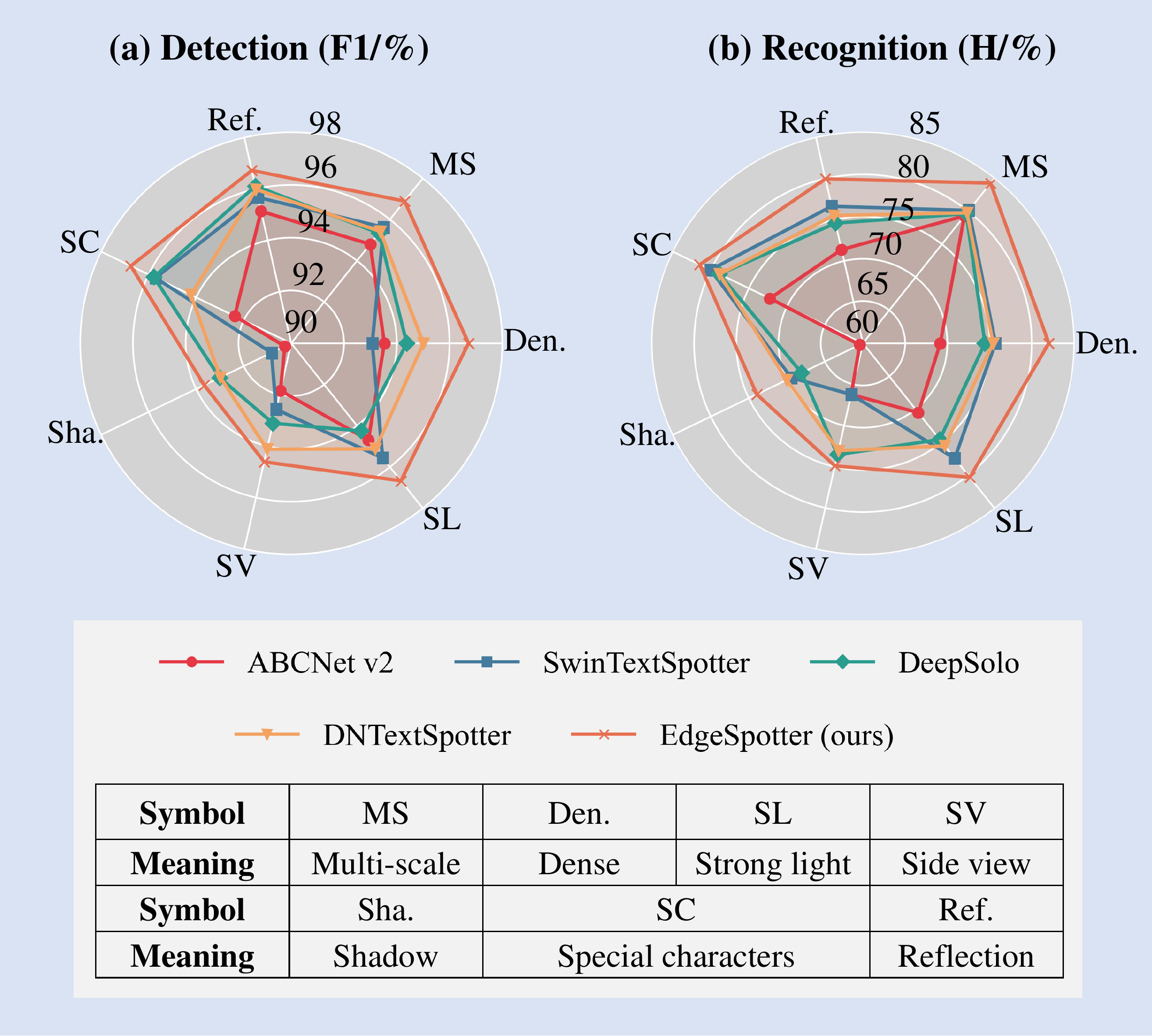}
    \setlength{\abovecaptionskip}{-5pt}
    \caption
    {
 The spotting accuracy of EdgeSpotter and other SOTA methods under
seven challenging attributes. Best viewed in color.
    }
    \label{fig:fig6}
    \vspace{-10pt}
\end{figure}
\begin{table}[!b]
\vspace{-5pt}
\caption{Ablation study on IPM. “EMT” and “FSCRS” denote efficient mixer Transformer and feature sampling with catmull-rom splines, respectively.  Red represents the best result.}
\vspace{-2pt}
\centering
\setlength{\tabcolsep}{2.0mm}
\renewcommand{\arraystretch}{1.1}
\begin{tabular}[h]{@{}l|c|c|c|c|c|c@{}}
\toprule[0.5mm]
\multirow{2}*{\textbf{Method}} & \multirow{2}*{\textbf{EMT}} & \multirow{2}*{\textbf{FSCRS}} & \multicolumn{3}{c|}{\textbf{Detection}} & \textbf{Recognition}\\
\cline{4-7}
 & & &\textbf{P($\%$)} & \textbf{R($\%$)} & \textbf{F1($\%$)} & H($\%$)\\
\midrule[0.25mm]
Baseline &   &  &93.10 &87.22  &90.06 &74.34\\
Baseline+ & & \Checkmark &97.18  & 93.86  & 95.49  & 79.04 \\
Baseline+& \Checkmark & & 97.26  &92.52  &94.83  &76.98 \\
Baseline+ & \Checkmark& \Checkmark &\textcolor{red}{98.16} & \textcolor{red}{94.52} &\textcolor{red}{96.31} &\textcolor{red}{81.25} \\
\bottomrule[0.5mm]
\end{tabular}
\label{tab:tab2}
\end{table}
\subsection{Ablation Studies}
In this section, the contributions of each module are analyzed through experiments conducted on \textbf{IPM}. The baseline model is defined as the network consisting only of feature extraction, Transformer encoding, and Transformer decoding. As shown in TABLE~\ref{tab:tab2}, the baseline network achieves an F1 score of \textbf{90.06\%}, and a recognition accuracy (H) of \textbf{74.34\%}, which is lower than all E2E Transformer-based methods compared in this study. After introducing FSCRS, F1 increases to \textbf{95.49\%}, a \textbf{5.43\%} improvement, and H rises to \textbf{79.04\%}, which is comparable to or even surpasses the performance of other SOTA methods. This demonstrates that FSCRS significantly enhances the feature extraction capabilities, leading to improved text spotting performance in industrial panels. Furthermore, the introduction of EMT alone results in a \textbf{4.77\%} increase in F1 and a \textbf{2.64\%} increase in H. Notably, EMT leads to a more significant improvement in text localization than recognition, highlighting its effectiveness in facilitating the deep interaction of spatial information across multi-level features, thus further enhancing the performance in text spotting across arbitrary scales.
 \subsection{Real-World Applications} 
As shown in Fig. \ref{fig:fig7}, to demonstrate the applicability of EdgeSpotter in real industrial panel monitoring, extensive tests were conducted on a self-built intelligent vision system equipped with an Intel \textit{i}9 CPU and NVIDIA RTX 3070 GPU. In the actual test, EdgeSpotter achieved a spotting speed of over \textbf{25 FPS} while maintaining accuracy. To further evaluate the quality of spotting, the average score for each test group was calculated. TEST 1 demonstrates that the proposed method effectively detects high-quality results in multi-scale text scenarios. TEST 2 focuses on evaluating the performance in dense text and bright light interference. TEST 3 presents a more challenging monitoring scenario. Despite these challenges, EdgeSpotter achieved outstanding spotting results, thanks to its robust ability to extract multi-scale and dense text features. In conclusion, EdgeSpotter can reliably identify text information in industrial panels under complex conditions, making it a valuable tool for monitoring real-world industrial panels.
\section{Conclusions}
This work proposes EdgeSpotter, a novel text spotter deployed in a lightweight, portable, low-power vision system for intelligent industrial panel monitoring. The goal is to reduce labor costs and enable full-cycle monitoring of industrial production. To address scale variations and ensure efficiency, EdgeSpotter introduces EMT to capture cross-scale information and establish inter-layer dependencies. FSCRS is used to encode information from dense texts. Experimental results show that EdgeSpotter effectively performs text spotting on complex industrial panels. In conclusion, we believe this work will significantly advance industrial automation.
\section*{Acknowledgment}
This work was supported by the National Natural Science Foundation of China (U24B20161).
%
%
\bibliographystyle{IEEEtran}
\normalem
\balance
\bibliography{EdgeSpotter}

\begin{thebibliography}{10}
\providecommand{\url}[1]{#1}
\csname url@samestyle\endcsname
\providecommand{\newblock}{\relax}
\providecommand{\bibinfo}[2]{#2}
\providecommand{\BIBentrySTDinterwordspacing}{\spaceskip=0pt\relax}
\providecommand{\BIBentryALTinterwordstretchfactor}{4}
\providecommand{\BIBentryALTinterwordspacing}{\spaceskip=\fontdimen2\font plus
\BIBentryALTinterwordstretchfactor\fontdimen3\font minus \fontdimen4\font\relax}
\providecommand{\BIBforeignlanguage}[2]{{%
\expandafter\ifx\csname l@#1\endcsname\relax
\typeout{** WARNING: IEEEtran.bst: No hyphenation pattern has been}%
\typeout{** loaded for the language `#1'. Using the pattern for}%
\typeout{** the default language instead.}%
\else
\language=\csname l@#1\endcsname
\fi
#2}}
\providecommand{\BIBdecl}{\relax}
\BIBdecl

\bibitem{gao2022review}
Y.~Gao, X.~Li, X.~V. Wang, L.~Wang, and L.~Gao, ``{A Review on Recent Advances in Vision-Based Defect Recognition towards Industrial Intelligence},'' \emph{Journal of Manufacturing Systems}, vol.~62, pp. 753--766, 2022.

\bibitem{shang2024methods}
Y.~Shang, H.~He, Z.~Zhang, X.~Zhang, Z.~Shen, J.~Yin, X.~Miao \emph{et~al.}, ``{Methods for Monitoring the Photovoltaic Panel: A Review},'' in \emph{Proceedings of the International Conference on Agro-Geoinformatics (Agro-Geoinformatics)}, 2024, pp. 1--5.

\bibitem{jiang2024comprehensive}
F.~Jiang, X.~Yuan, L.~Hu, G.~Xie, Z.~Zhang, X.~Li, J.~Hu, C.~Wang, and H.~Wang, ``{A Comprehensive Review of Energy Storage Technology Development and Application for Pure Electric Vehicles},'' \emph{Journal of Energy Storage}, vol.~86, p. 111159, 2024.

\bibitem{deng2023end}
X.~Deng, X.~Chen, D.~Cao, K.~Ren, and P.~Z. Sun, ``{An End-to-End Tag Recognition Architecture for Industrial Meter},'' \emph{IEEE Transactions on Industrial Informatics}, vol.~20, no.~1, pp. 117--126, 2023.

\bibitem{park2024ocr}
C.-W. Park, V.~Palakonda, S.~Yun, I.-M. Kim, and J.-M. Kang, ``{OCR-Diff: A Two-Stage Deep Learning Framework for Optical Character Recognition Using Diffusion Model in Industrial Internet-of-Things},'' \emph{IEEE Internet of Things Journal}, vol.~11, no.~15, pp. 25\,997--26\,000, 2024.

\bibitem{wang2021pan++}
W.~Wang, E.~Xie, X.~Li, X.~Liu, D.~Liang, Z.~Yang, T.~Lu, and C.~Shen, ``{PAN++: Towards Efficient and Accurate End-to-End Spotting of Arbitrarily-Shaped Text},'' \emph{IEEE Transactions on Pattern Analysis and Machine Intelligence}, vol.~44, no.~9, pp. 5349--5367, 2021.

\bibitem{liu2020abcnet}
Y.~Liu, H.~Chen, C.~Shen, T.~He, L.~Jin, and L.~Wang, ``{ABCNet: Real-Time Scene Text Spotting with Adaptive Bezier-Curve Network},'' in \emph{Proceedings of the IEEE/CVF Conference on Computer Vision and Pattern Recognition (CVPR)}, 2020, pp. 9809--9818.

\bibitem{liu2021abcnet}
Y.~Liu, C.~Shen, L.~Jin, T.~He, P.~Chen, C.~Liu, and H.~Chen, ``{ABCNet v2: Adaptive Bezier-Curve Network for Real-Time End-to-End Text Spotting},'' \emph{IEEE Transactions on Pattern Analysis and Machine Intelligence}, vol.~44, no.~11, pp. 8048--8064, 2021.

\bibitem{vaswani2017attention}
A.~Vaswani, N.~Shazeer, N.~Parmar, J.~Uszkoreit, L.~Jones, A.~N. Gomez, {\L}.~Kaiser, and I.~Polosukhin, ``{Attention Is All You Need},'' in \emph{Proceedings of Neural Information Processing Systems (NIPS)}, vol.~30, 2017, pp. 20\,750–--20\,762.

\bibitem{han2022survey}
K.~Han, Y.~Wang, H.~Chen, X.~Chen, J.~Guo, Z.~Liu, Y.~Tang, A.~Xiao, C.~Xu, Y.~Xu \emph{et~al.}, ``{A Survey on Vision Transformer},'' \emph{IEEE Transactions on Pattern Analysis and Machine Intelligence}, vol.~45, no.~1, pp. 87--110, 2022.

\bibitem{huang2022swintextspotter}
M.~Huang, Y.~Liu, Z.~Peng, C.~Liu, D.~Lin, S.~Zhu, N.~Yuan, K.~Ding, and L.~Jin, ``{Swintextspotter: Scene Text Spotting via Better Synergy between Text Detection and Text Recognition},'' in \emph{Proceedings of the IEEE/CVF Conference on Computer Vision and Pattern Recognition (CVPR)}, 2022, pp. 4593--4603.

\bibitem{ye2023deepsolo}
M.~Ye, J.~Zhang, S.~Zhao, J.~Liu, T.~Liu, B.~Du, and D.~Tao, ``{Deepsolo: Let Transformer Decoder with Explicit Points Solo for Text Spotting},'' in \emph{Proceedings of the IEEE/CVF Conference on Computer Vision and Pattern Recognition (CVPR)}, 2023, pp. 19\,348--19\,357.

\bibitem{qiao2024dntextspotter}
Q.~Qiao, Y.~Xie, J.~Gao, T.~Wu, S.~Huang, J.~Fan, Z.~Cao, Z.~Wang, and Y.~Zhang, ``{DNTextSpotter: Arbitrary-Shaped Scene Text Spotting via Improved Denoising Training},'' in \emph{Proceedings of the ACM International Conference on Multimedia (ACM MM)}, 2024, pp. 10\,134--10\,143.

\bibitem{ren2016faster}
S.~Ren, K.~He, R.~Girshick, and J.~Sun, ``{Faster R-CNN: Towards Real-Time Object Detection with Region Proposal Networks},'' \emph{IEEE Transactions on Pattern Analysis and Machine Intelligence}, vol.~39, no.~6, pp. 1137--1149, 2016.

\bibitem{zhou2017east}
X.~Zhou, C.~Yao, H.~Wen, Y.~Wang, S.~Zhou, W.~He, and J.~Liang, ``{EAST: an Efficient and Accurate Scene Text Detector},'' in \emph{Proceedings of the IEEE conference on Computer Vision and Pattern Recognition (CVPR)}, 2017, pp. 5551--5560.

\bibitem{liao2020real}
M.~Liao, Z.~Wan, C.~Yao, K.~Chen, and X.~Bai, ``{Real-Time Scene Text Detection with Differentiable Binarization},'' in \emph{Proceedings of the AAAI Conference on Artificial Intelligence (AAAI)}, vol.~34, no.~07, 2020, pp. 11\,474--11\,481.

\bibitem{cheng2018aon}
Z.~Cheng, Y.~Xu, F.~Bai, Y.~Niu, S.~Pu, and S.~Zhou, ``{AON: Towards Arbitrarily-Oriented Text Recognition},'' in \emph{Proceedings of the IEEE Conference on Computer Vision and Pattern Recognition (CVPR)}, 2018, pp. 5571--5579.

\bibitem{li2017towards}
H.~Li, P.~Wang, and C.~Shen, ``{Towards End-to-End Text Spotting with Convolutional Recurrent Neural Networks},'' in \emph{Proceedings of the IEEE International Conference on Computer Vision (ICCV)}, 2017, pp. 5238--5246.

\bibitem{he2018end}
T.~He, Z.~Tian, W.~Huang, C.~Shen, Y.~Qiao, and C.~Sun, ``{An End-to-End Textspotter with Explicit Alignment and Attention},'' in \emph{Proceedings of the IEEE Conference on Computer Vision and Pattern Recognition (CVPR)}, 2018, pp. 5020--5029.

\bibitem{lyu2018mask}
P.~Lyu, M.~Liao, C.~Yao, W.~Wu, and X.~Bai, ``{Mask Textspotter: An End-to-End Trainable Neural Network for Spotting Text with Arbitrary Shapes},'' in \emph{Proceedings of the European Conference on Computer Vision (ECCV)}, 2018, pp. 67--83.

\bibitem{feng2019textdragon}
W.~Feng, W.~He, F.~Yin, X.-Y. Zhang, and C.-L. Liu, ``{Textdragon: An End-to-End Framework for Arbitrary Shaped Text Spotting},'' in \emph{Proceedings of the IEEE/CVF International Conference on Computer Vision (ICCV)}, 2019, pp. 9076--9085.

\bibitem{liao2020mask}
M.~Liao, G.~Pang, J.~Huang, T.~Hassner, and X.~Bai, ``{Mask TextSpotter v3: Segmentation Proposal Network for Robust Scene Text Spotting},'' in \emph{Proceedings of the European Conference on Computer Vision (ECCV)}, 2020, pp. 706--722.

\bibitem{qiao2021mango}
L.~Qiao, Y.~Chen, Z.~Cheng, Y.~Xu, Y.~Niu, S.~Pu, and F.~Wu, ``{Mango: A Mask Attention Guided One-Stage Scene Text Spotter},'' in \emph{Proceedings of the AAAI Conference on Artificial Intelligence (AAAI)}, vol.~35, no.~3, 2021, pp. 2467--2476.

\bibitem{zhang2022text}
X.~Zhang, Y.~Su, S.~Tripathi, and Z.~Tu, ``{Text Spotting Transformers},'' in \emph{Proceedings of the IEEE/CVF Conference on Computer Vision and Pattern Recognition (CVPR)}, 2022, pp. 9519--9528.

\bibitem{girshick2015fast}
R.~Girshick, ``{Fast R-CNN},'' in \emph{Proceedings of the IEEE International Conference on Computer Vision (ICCV)}, 2015, pp. 1440--1448.

\bibitem{litman2020scatter}
R.~Litman, O.~Anschel, S.~Tsiper, R.~Litman, S.~Mazor, and R.~Manmatha, ``{Scatter: Selective Context Attentional Scene Text Recognizer},'' in \emph{Proceedings of the IEEE/CVF Conference on Computer Vision and Pattern Recognition (CVPR)}, 2020, pp. 11\,962--11\,972.

\bibitem{he2017mask}
K.~He, G.~Gkioxari, P.~Doll{\'a}r, and R.~Girshick, ``{Mask R-CNN},'' in \emph{Proceedings of the IEEE International Conference on Computer Vision (ICCV)}, 2017, pp. 2961--2969.

\bibitem{ronen2022glass}
R.~Ronen, S.~Tsiper, O.~Anschel, I.~Lavi, A.~Markovitz, and R.~Manmatha, ``{Glass: Global to Local Attention for Scene-Text Spotting},'' in \emph{Procedings of the European Conference on Computer Vision (ECCV)}, 2022, pp. 249--266.

\bibitem{shaker2023swiftformer}
A.~Shaker, M.~Maaz, H.~Rasheed, S.~Khan, M.-H. Yang, and F.~S. Khan, ``{Swiftformer: Efficient Additive Attention for Transformer-based Real-Time Mobile Vision Applications},'' in \emph{Proceedings of the IEEE/CVF International Conference on Computer Vision (ICCV)}, 2023, pp. 17\,425--17\,436.

\bibitem{twigg2003catmull}
C.~Twigg, ``{Catmull-Rom Splines},'' \emph{Computer}, vol.~41, no.~6, pp. 4--6, 2003.

\bibitem{wang2021pgnet}
P.~Wang, C.~Zhang, F.~Qi, S.~Liu, X.~Zhang, P.~Lyu, J.~Han, J.~Liu, E.~Ding, and G.~Shi, ``{PGNet: Real-Time Arbitrarily-Shaped Text Spotting with Point Gathering Network},'' in \emph{Proceedings of the AAAI Conference on Artificial Intelligence (AAAI)}, vol.~35, no.~4, 2021, pp. 2782--2790.

\bibitem{karatzas2015icdar}
D.~Karatzas, L.~Gomez-Bigorda, A.~Nicolaou, S.~Ghosh, A.~Bagdanov, M.~Iwamura, J.~Matas, L.~Neumann, V.~R. Chandrasekhar, S.~Lu \emph{et~al.}, ``{ICDAR 2015 Competition on Robust Reading},'' in \emph{Proceedings of the International Conference on Document Analysis and Recognition (ICDAR)}, 2015, pp. 1156--1160.

\bibitem{liu2019curved}
Y.~Liu, L.~Jin, S.~Zhang, C.~Luo, and S.~Zhang, ``{Curved Scene Text Detection via Transverse and Longitudinal Sequence Connection},'' \emph{Pattern Recognition}, vol.~90, pp. 337--345, 2019.

\end{thebibliography}
\end{document}